\documentclass{article} 
\RequirePackage{amsmath,amsthm,amsfonts,amssymb}
\usepackage{iclr2017_conference,times}
\usepackage{hyperref}
\usepackage{url}
\usepackage{graphicx} 
\usepackage{algorithm}
\usepackage{algorithmic}
\usepackage{hyperref}

\title{Learning Identity Mappings with Residual Gates}

\author{Pedro H. P. ~Savarese\\
COPPE/PESC\\
Federal University of Rio de Janeiro\\
Rio de Janeiro, Brazil \\
\texttt{savarese@land.ufrj.br} \\
}

\author{Pedro H. P. ~Savarese \\
COPPE/PESC\\
Federal University of Rio de Janeiro\\
Rio de Janeiro, Brazil  \\
\texttt{savarese@land.ufrj.br} \\
\And
Leonardo O. ~Mazza \\
Poli \\
Federal University of Rio de Janeiro \\
Rio de Janeiro, Brazil \\
\texttt{leonardomazza@poli.ufrj.br} \\
\And
Daniel R. ~Figueiredo \\
COPPE/PESC \\
Federal University of Rio de Janeiro \\
Rio de Janeiro, Brazil \\
\texttt{daniel@land.ufrj.br} \\
}

\begin{document}

\maketitle

\begin{abstract}

We propose a new layer design by adding a linear gating mechanism to shortcut connections. By using a scalar parameter to control each gate, we provide a way to learn identity mappings by optimizing only one parameter. We build upon the motivation behind Residual Networks, where a layer is reformulated in order to make learning identity mappings less problematic to the optimizer. The augmentation introduces only one extra parameter per layer, and provides easier optimization by making degeneration into identity mappings simpler. We propose a new model, the Gated Residual Network, which is the result when augmenting Residual Networks. Experimental results show that augmenting layers provides better optimization, increased performance, and more layer independence. We evaluate our method on MNIST using fully-connected networks, showing empirical indications that our augmentation facilitates the optimization of deep models, and that it provides high tolerance to full layer removal: the model retains over $90 \%$ of its performance even after half of its layers have been randomly removed. We also evaluate our model on CIFAR-10 and CIFAR-100 using Wide Gated ResNets, achieving $3.65 \%$ and $18.27 \%$ error, respectively.

\end{abstract}

\section{Introduction}
\label{introduction}

As the number of layers of neural networks increase, effectively training its parameters becomes a fundamental problem (\cite{deephard}). Many obstacles challenge the training of neural networks, including vanishing/exploding gradients (\cite{hardtrain}), saturating activation functions (\cite{saturation}) and poor weight initialization (\cite{glorot}). Techniques such as unsupervised pre-training (\cite{aes}), non-saturating activation functions (\cite{relu}) and normalization (\cite{bn}) target these issues and enable the training of deeper networks. However, stacking more than a dozen layers still lead to a hard to train model. 

Recently, models such as Residual Networks (\cite{resnet1}) and Highway Neural Networks (\cite{highway}) permitted the design of networks with hundreds of layers. A key idea of these models is to allow for information to flow more freely through the layers, by using shortcut connections between the layer's input and output. This layer design greatly facilitates training, due to shorter paths between the lower layers and the network's error function. In particular, these models can more easily learn identity mappings in the layers, thus allowing the network to be deeper and learn more abstract representations (\cite{representations}). Such networks have been highly successful in many computer vision tasks. 

On the theoretical side, it is suggested that depth contributes exponentially more to the representational capacity of networks than width (\cite{exp1} \cite{exp2} \cite{exp3} \cite{exp4}). This agrees with the increasing depth of winning architectures on challenges such as ImageNet (\cite{resnet1} \cite{googlenet}).

Increasing the depth of networks significantly increases its representational capacity and consequently its performance, an observation supported by theory (\cite{exp1} \cite{exp2} \cite{exp3} \cite{exp4}) and practice (\cite{resnet1} \cite{googlenet}). Moreover, \cite{resnet1} showed that, by construction, one can increase a network's depth while preserving its performance. These two observations suggest that it suffices to stack more layers to a network in order to increase its performance. However, this behavior is not observed in practice even with recently proposed models, in part due to the challenge of training ever deeper networks.

\begin{figure}[!ht]
  \centering
    \includegraphics[width=0.7\textwidth]{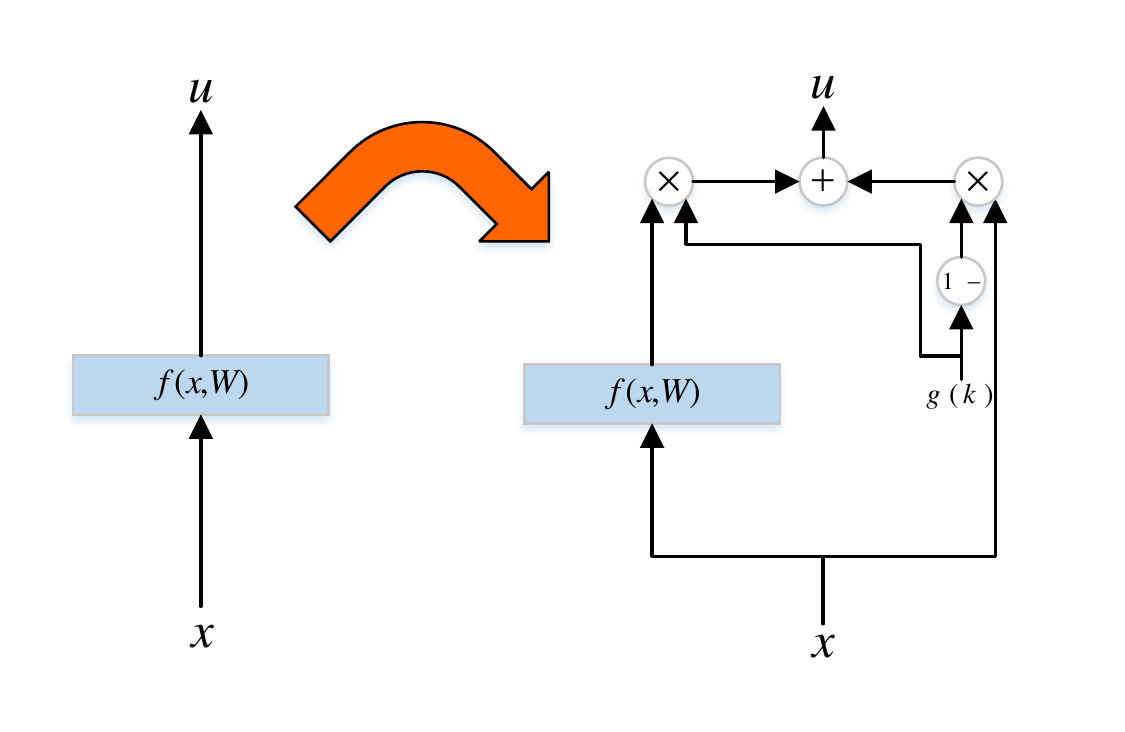}
  \caption{Gating mechanism applied to the shortcut connection of a layer. The key difference with Highway Networks is that only a scalar $k$ is used to regulate the gates instead of a tensor.}
\label{aug}
\end{figure}

In this work we aim to improve the training of deep networks by proposing a layer design that builds on Residual Networks and Highway Neural Networks. The key idea is to facilitate the learning of identity mappings by introducing a {\em gating mechanism} to the shortcut connection, as illustrated in Figure~\ref{aug}. Note that the shortcut connection is controlled by a gate that is parameterized with a scalar, $k$. This is a key difference from Highway Networks, where a tensor is used to regulate the shortcut connection, along with the incoming data. The idea of using a scalar is simple: it is easier to learn $k=0$ than to learn $W_g=0$ for a weight tensor $W_g$ controlling the gate. Indeed, this single scalar allows for stronger supervision on lower layers, by making gradients flow more smoothly in the optimization.

\begin{figure}[!ht]
  \centering
    \includegraphics[width=0.85\textwidth]{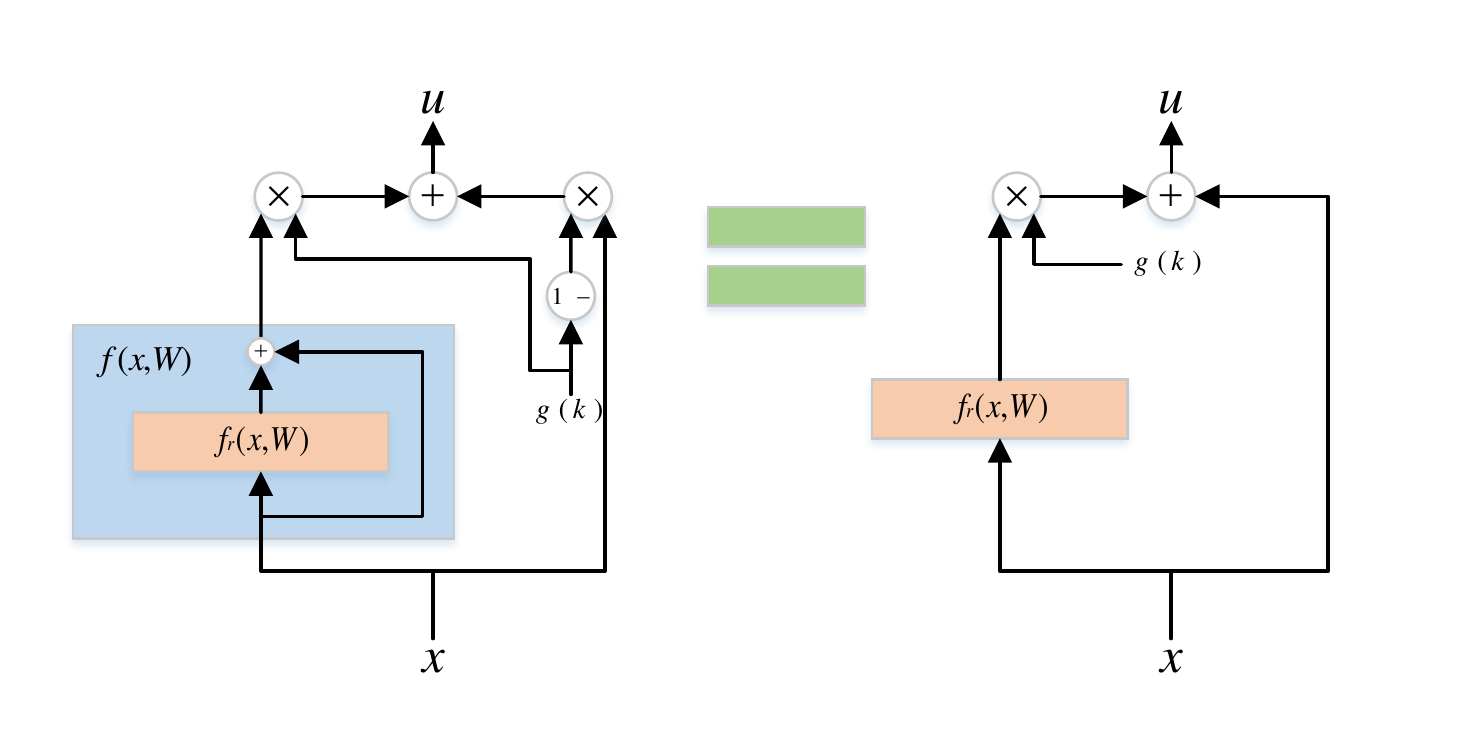}
  \caption{Proposed network design applied to Residual Networks. Note that the joint network design results in a shortcut path where the input remains unchanged. In this case, $g(k)$ can be interpreted as an amplifier or suppressor for the residual $f_r(x,W)$.}
\label{resaug}
\end{figure}



We apply our proposed network design to Residual Networks, as illustrated in Figure~\ref{resaug}. Note that in this case the layer becomes simply $u = g(k) f_r(x,W) + x$, where $f_r$ denotes the layer's residual function. Thus, the shortcut connection allows the input to flow freely without any interference of $g(k)$ through the layer. We will call this model Gated Residual Network, or GResNets. Again, note that learning identity mappings is again much easier in comparison to the original ResNets.

Note that layers that degenerated into identity mappings have no impact in the signal propagating through the network, and thus can be removed without affecting performance. The removal of such layers can be seen as a transposed application of sparse encoding (\cite{sparse}): transposing the sparsity from neurons to layers provides a form to prune them entirely from the network. Indeed, we show that performance decays slowly in GResNets when layers are removed, when compared to ResNets.

We evaluate the performance of the proposed design in two experiments. First, we evaluate fully-connected GResNets on MNIST and compare it with fully-connected ResNets, showing superior performance and robustness to layer removal. Second, we apply our model to Wide ResNets (\cite{wide}) and test its performance on CIFAR, obtaining results that are superior to all previously published results (to the best of our knowledge). These findings indicate that learning identity mappings is a fundamental aspect of learning in deep networks, and designing models where this is easier seems highly effective.

\section{Augmentation with Residual Gates}

\subsection{Theoretical Intuition}

Recall that a network's depth can always be increased without affecting its performance -- it suffices to add layers that perform identity mappings. Consider a classic fully-connected ReLU network with layers defined as $u = ReLU( \langle x,W \rangle )$. When adding a new layer, if we initialize $W$ to the identity matrix $I$, we have:
\begin{equation*}
	u = ReLU(\langle x, I \rangle) = ReLU(x) = x
\end{equation*}
The last step holds since $x$ is an output of a previous ReLU layer, and $ReLU(ReLU(x)) = ReLU(x)$. Thus, adding more layers should only improve performance. However, how can a network with more layers learn to yield performance superior than a network with less layers? A key observation is that if learning identity mapping is easy, then the network with more layers is more likely to yield superior performance, as it can more easily recover the performance of a smaller network through identity mappings.

\begin{figure}[!ht]
  \centering
    \includegraphics[width=0.9\textwidth]{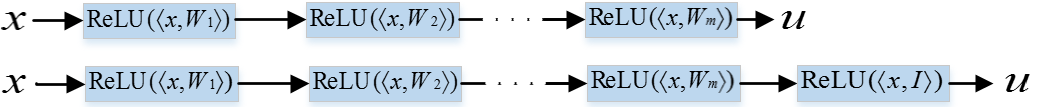}
  \caption{A network can have layers added to it without losing performance. Initially, a network has $m$ ReLU layers with parameters $\{W_1, \dots, W_m \}$. A new, $(m+1)$-th layer is added with $W_{m+1} = I$. This new layer will perform an identity mapping, therefore the two models are equivalent.}
\label{id}
\end{figure}

The layer design of Residual Networks allows for deeper models to be trained due to its shortcut connections. Note that in ResNets the identity mapping is learned when $W = 0$ instead of $W = I$. Considering a residual layer $u = ReLU( \langle x,W \rangle ) + x$, we have:
\begin{equation*}
	u = ReLU(\langle x, 0 \rangle) + x = ReLU(0) + x = x
\end{equation*}
Intuitively, residual layers can degenerate into identity mappings more effectively since learning an all-zero matrix is easier than learning the identity matrix. To support this argument, consider weight parameters randomly initialized with zero mean. Hence, the point $W = 0$ is located exactly in the center of the probability mass distribution used to initialize the weights.

However, assuming that residual layers can trivially learn the parameter set $W = 0$ implies ignoring the randomness when initializing the weights. We demonstrate this by calculating the expected component-wise distance between $W_{o}$ and the origin. Here, $W_{o}$ denotes the weight tensor after initialization and prior to any optimization. Note that the distance between $W_{o}$ and the origin captures the effort for a network to learn identity mappings:

\begin{equation*}
	E \Big [  (W_{o} - 0)^2  \Big ] = E \Big [  W_{o}^2  \Big ] =  Var \Big [  W_{o}  \Big ]
\end{equation*}

Note that the distance is given by the distribution's variance, and there is no reason to assume it to be negligible. Additionally, the fact that Residual Networks still suffer from optimization issues caused by depth (\cite{stdepth}) further supports this claim. 

Some initialization schemes propose a variance in the order of $O(\frac{1}{n})$ (\cite{glorotinit}, \cite{prelu}), however this represents the distance for each individual parameter in $W$. For tensors with $O(n^2)$ parameters, the total distance --  either absolute or Euclidean --  between $W_{o}$ and the origin will be in the order of $O(n)$.

\subsection{Residual Gates}

As previously mentioned, the key contribution in this work is the proposal of a layer design where learning a single scalar parameter suffices in order for the layer to degenerate into an identity mapping. As in Highway Networks, we propose the addition of gated shortcut connections. Our gates, however, are parametrized by a single scalar value, being easier to analyze and learn. In our model, the effort required to learn identity mappings does not depend on any parameter, such as the layer width, in sharp contrast to prior models.

Our design is as follows: a layer $u = f(x,W)$ becomes $u = g(k) f(x,W) + (1 - g(k)) x$, where $k$ is a scalar parameter. This design is illustrated in Figure~\ref{aug}. Note that such layer can quickly degenerate by setting $g(k)$ to $0$. Using the ReLU activation function as $g$, it suffices that $k \leq 0$ for $g(k) = 0$.

By adding an extra parameter, the dimensionality of the cost surface also grows by one. This new dimension, however, can be easily understood due to the specific nature of the layer reformulation. The original surface is maintained on the $k = 1$ slice, since the gated model becomes equivalent to the original one. On the $k = 0$ slice we have an identity mapping, and the associated cost for all points in such slice is the same cost associated with the point $\{k = 1, W = I\}$: this follows since both parameter configurations correspond to identity mappings, therefore being equivalent. Lastly, due to the linear nature of  $g(k)$ and consequently of the gates, all other slices $k \neq 0, k \neq 1$ will be a linear combination between the slices $k = 0$ and $k = 1$.

We proceed to use residual layers as the basis for our design, for two reasons. First, they are the current standard for computer vision tasks. Second, ResNets lack means to regulate the residuals, therefore a linear gating mechanism might not only allow deeper models, but could also improve performance. Thus, the residual layer is given by:

\begin{equation*}
	u = f(x,W) = f_r(x, W) + x
\end{equation*}

where $f_r(x,W)$ is the layer's residual function -- in our case, \textbf{BN-ReLU-Conv-BN-ReLU-Conv}. Our approach changes this layer by adding a linear gate, yielding:

\begin{align*}
\begin{split}
	 u &= g(k) f(x,W) + (1 - g(k))x \\
	&=  g(k) ( f_r(x, W) + x ) + (1 - g(k))x \\
	&= g(k) f_r(x,W) + x
\end{split}
\end{align*}

Our approach applied to residual layers is shown in Figure \ref{resaug}. The resulting layer maintains the shortcut connection unaltered, which according to \cite{resnet2} is a desired property when designing residual blocks. As $(1 - g(k))$ vanishes from the formulation, $g(k)$ stops acting as a dual gating mechanism and can be interpreted as a flow regulator. Note that this model introduces a single scalar parameter per layer block. This new dimension can be interpreted as discussed above, except that the slice $k = 0$ is equivalent to $\{k = 1, W = 0\}$, since an identity mapping is learned when $W = 0$ in ResNets.

\section{Experiments}

All models were implemented on Keras (\cite{keras}) or on Torch (\cite{t7}), and were executed on a Geforce GTX 1070. Larger models or more complex datasets, such as the ImageNet (\cite{imagenet}), were not explored due to hardware limitations.

\subsection{MNIST}

The MNIST dataset (\cite{mnist}) is composed of $60,000$ greyscale images with $28 \times 28$ pixels. Images represent handwritten digits, resulting in a total of 10 classes. We trained three types of fully-connected models: classical plain networks, ResNets and GResNets.

The networks consist of a linear layer with 50 neurons, followed by $d$ layers with 50 neurons each, and lastly a softmax layer for classification. Only the $d$ middle layers differ between the three architectures -- the first linear layer and the softmax layer are the same in all experiments.

For plain networks, each layer performs dot product, followed by Batch Normalization and a ReLU activation function.

Initial tests with pre-activations (\cite{resnet2}) resulted in poor performance on the validation set, therefore we opted for the traditional \textbf{Dot-BN-ReLU} layer when designing Residual Networks. Each residual block is consists of two layers, as conventional.

All networks were trained using Adam (\cite{adam}) with Nesterov momentum (\cite{adamnest}) for a total of 100 epochs using mini-batches of size 128. No learning rate decay was used: we kept the learning rate and momentum fixed to $0.002$ and $0.9$ during the whole training.

For preprocessing, we divided each pixel value by 255, normalizing their values to $[0,1]$.

\begin{figure}[!ht]
  \centering
    \includegraphics[width=1.0\textwidth]{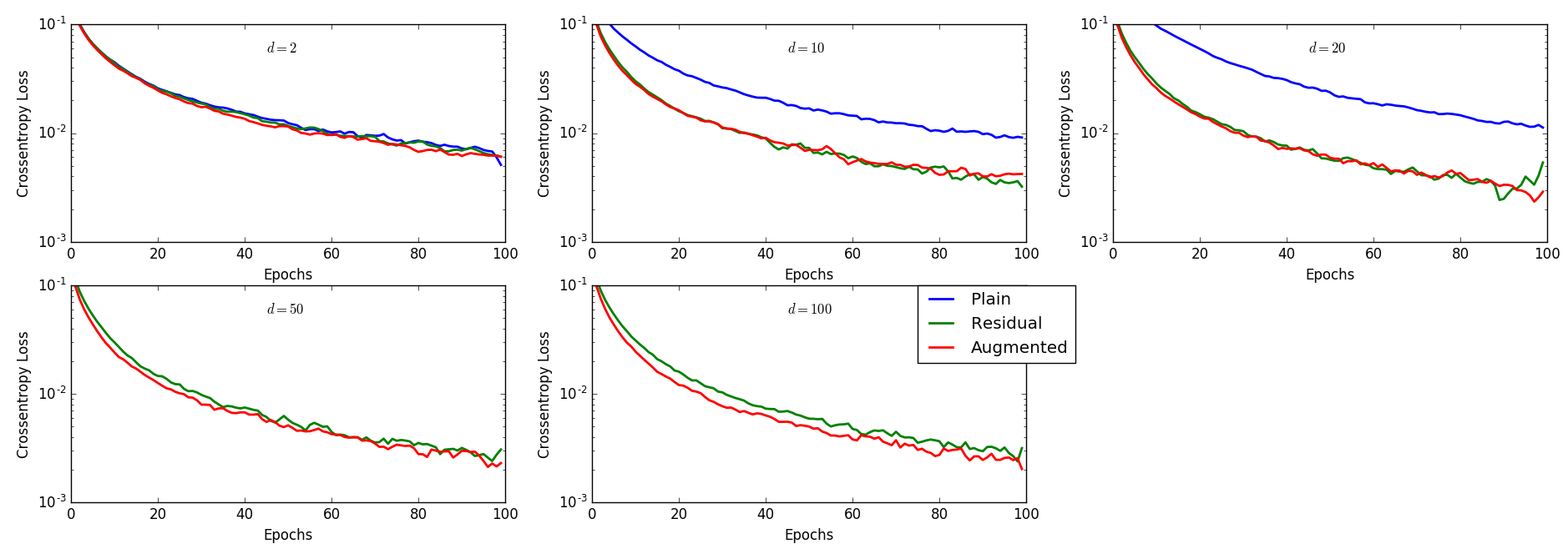}
  \caption{Train loss for plain (classical), residual and gated residual networks (GResNet), with $d = \{2,10,20,50,100\}$. As the models get deeper, the error reduction due to the augmentation increases.}
\label{mnist_loss}
\end{figure}

The training curves for classical plain networks, ResNets and GResNets with varying depth are shown in Figure \ref{mnist_loss}. The distance between the curves increase with the depth, showing that the augmentation helps the training of deeper models.

\begin{table}[h!]
\centering
    \begin{tabular}{ | l | c | c | c |}
    \hline
    Depth = $d+2$ & Classical & ResNet & GResNet \\ \hline
    $d=2$  & 			2.29 & 	2.20 & 	2.17 \\ \hline
    $d=10$ & 			2.22 & 	1.64 & 	1.60 \\ \hline
    $d=20$ & 			2.21 & 	1.61 & 	1.57 \\ \hline
    $d=50$ & 			60.37 & 	1.62 & 	1.48 	\\ \hline
    $d=100$ &		90.20 & 	1.50 & 	1.26 	\\ \hline
    \end{tabular}
\caption{Test error (\%) on the MNIST dataset for fully-connected networks. GResNets achieve lower error than ResNets in all experiments. Classical fully-connected networks perform worse and fail to converge for $d = 50$ and $d = 100$.}
    \label{mnist_table}   
\end{table}

Table \ref{mnist_table} shows the test error for each depth and architecture. ResNets converge in experiments with $d = 50$ and $d = 100$ ($52$ and $102$ layers, respectively), while classical models do not. 

Gated Residual Networks perform better in all settings, and the performance boost is more noticeable with increased depths. The relative error decreased approximately $2.5 \%$ for $d = \{2,10,20\}$, $8.7 \%$ for $d=50$ and $16\%$ for $d = 100$.

\begin{table}[h!]
\centering
    \begin{tabular}{ | l | l |}
    \hline
    Depth = $d+2$ & Mean $k$ \\ \hline
    $d=2 $ & 			5.58 	\\ \hline
    $d=10$ & 			2.54	\\ \hline
    $d=20$ & 			1.73	\\ \hline
    $d=50$ & 			1.04 	\\ \hline
    $d=100$ & 			0.67 	\\ \hline
    \end{tabular}
\caption{Mean $k$ for increasingly deep Gated Residual Networks.}
    \label{mnist_k}   
\end{table}

As observed in Table \ref{mnist_k}, the mean values of $k$ decrease as the model gets deeper, showing that shortcut connections have less impact on shallow networks. This agrees with empirical results that ResNets perform better than classical plain networks as the depth increases.

\begin{figure}[!ht]
  \centering
    \includegraphics[width=1\textwidth]{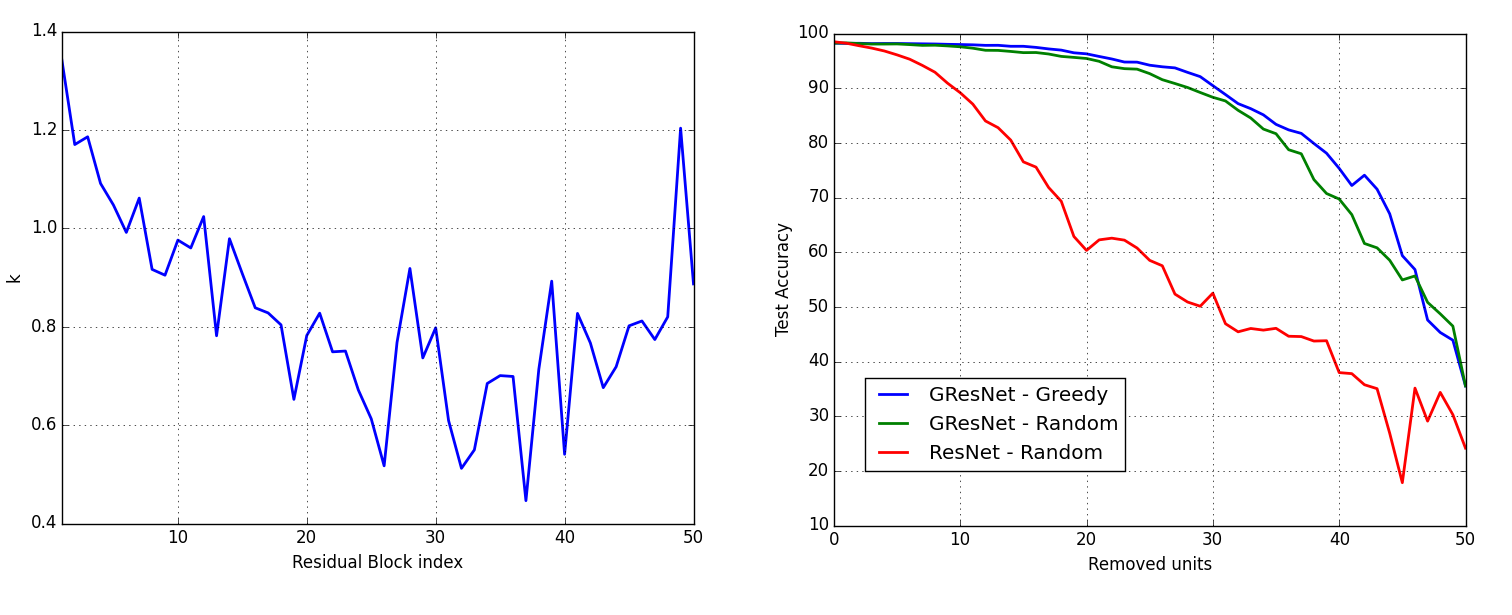}
  \caption{\textit{Left: } Values for $k$ according to ascending order of residual blocks. The first block, consisted of the first two layers of the network, has index 1, while the last block -- right before the softmax layer -- has index 50. \textit{Right:} Test accuracy (\%) according to the number of removed layers. Gated Residual Networks are more robust to layer removal, and maintain decent results even after half of the layers have been removed. }
\label{pruning}
\end{figure}

We also analyzed how layer removal affects ResNets and GResNets. We compared how the deepest networks ($d = 100$) behave as residual blocks composed of 2 layers are completely removed from the models. The final values for each $k$ parameter, according to its corresponding residual block, is shown in Figure \ref{pruning}. We can observe that layers close to the middle of the network have a smaller $k$ than these in the beginning or the end. Therefore, the middle layers have less importance by due to being closer to identity mappings.

Results are shown in Figure \ref{pruning}. For Gated Residual Networks, we prune pairs of layers following two strategies. One consists of pruning layers in a greedy fashion, where blocks with the smallest $k$ are removed first. In the other we remove blocks randomly. We present results using both strategies for GResNets, and only random pruning for ResNets since they lack the $k$ parameter.

The greedy strategy is slightly better for Gated Residual Networks, showing that the $k$ parameter is indeed a good indicator of a layer's importance for the model, but that layers tend to assume the same level of significance. In a fair comparison, where both models are pruned randomly, GResNets retain a satisfactory performance even after half of its layers have been removed, while ResNets suffer performance decrease after just a few layers.

Therefore augmented models are not only more robust to layer removal, but can have a fair share of their layers pruned and still perform well. Faster predictions can be generated by using a pruned version of an original model.

\subsection{CIFAR}

The CIFAR datasets (\cite{cifar}) consists of $60,000$ color images with $32 \times 32$ pixels each. CIFAR-10 has a total of 10 classes, including pictures of cats, birds and airplanes. The CIFAR-100 dataset is composed of the same number of images, however with a total of 100 classes.

Residual Networks have surpassed state-of-the-art results on CIFAR. We test GResNets, Wide GResNets (\cite{wide}) and compare them with their original, non-augmented models.

For pre-activation ResNets, as described in \cite{resnet2}, we follow the original implementation details. We set an initial learning rate of 0.1, and decrease it by a factor of 10 after 50\% and 75\% epochs. SGD with Nesterov momentum of 0.9 are used for optimization, and the only pre-processing consists of mean subtraction. Weight decay of 0.0001 is used for regularization, and Batch Normalization's momentum is set to 0.9.

We follow the implementation from \cite{wide} for Wide ResNets. The learning rate is initialized as 0.1, and decreases by a factor of 5 after 30\%, 60\% and 80\% epochs. Images are mean/std normalized, and a weight decay of 0.0005 is used for regularization. When dropout is specified, we apply 0.3 dropout (\cite{dropout}) between convolutions. All other details are the same as for ResNets.

\begin{figure}[!ht]
  \centering
    \includegraphics[width=1\textwidth]{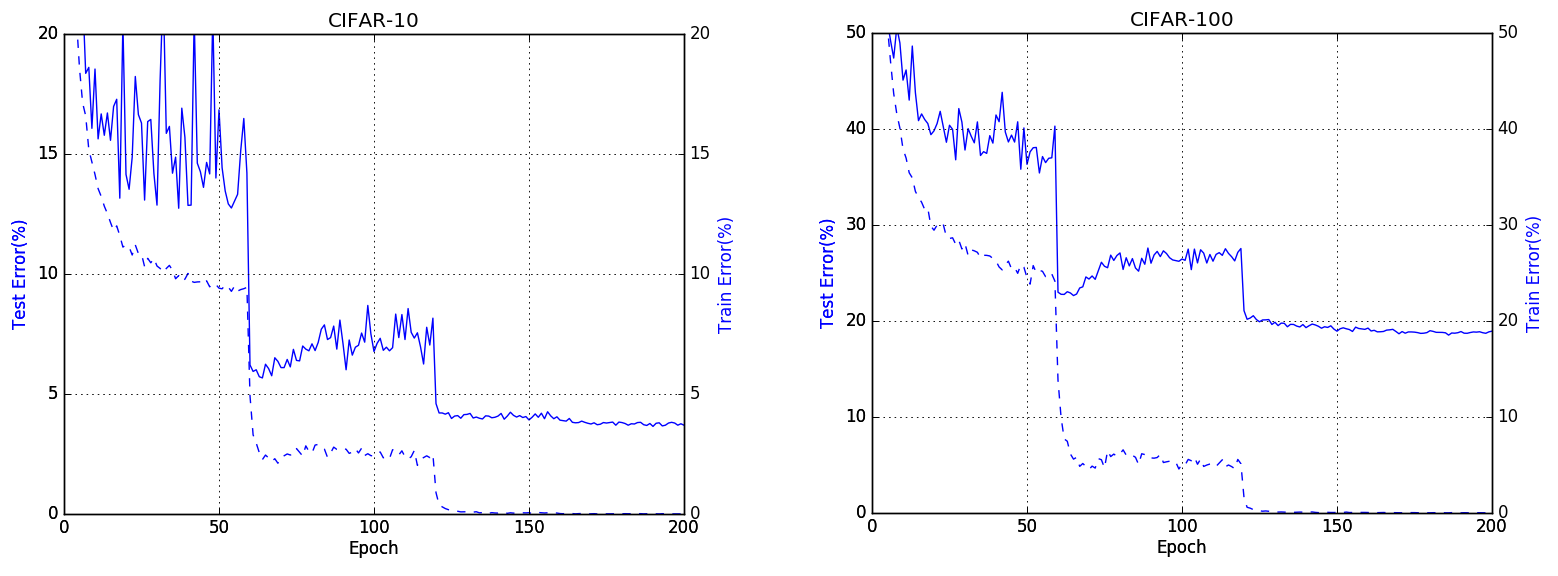}
  \caption{Training and test curves, showing error (\%) on training and test sets. Dashed lines represent training error, whereas solid lines represent test error.}
\label{cifar}
\end{figure}

For both architectures we use moderate data augmentation: images are padded with 4 pixels, and we take random  crops of size $32 \times 32$ during training. Additionally, each image is horizontally flipped with  $50\%$  probability. We use batch size 128 for all experiments.

For all gated networks, we initialize $k$ with a constant value of $1$. One crucial question is whether weight decay should be applied to the $k$ parameters. We call this "$k$ decay", and also compare GResNets and Wide GResNets when it is applied with the same magnitude of the weight decay: 0.0001 for GResNet and 0.0005 for Wide GResNet.

\begin{table}[h!]
\centering
    \begin{tabular}{ | l | c | c | c |}
    \hline
    Acc    & 		  Original & Gated & Gated ($k$ decay)  \\ \hline
    Resnet 5 & 		7.16 & 	\textbf{6.67} &    7.04\\ \hline
    Wide ResNet (4,10) + Dropout &  3.89 & 	\textbf{3.65}   & 3.74 \\ \hline
    \end{tabular}
\caption{Test error (\%) on the CIFAR-10 dataset, for ResNets, Wide ResNets and their augmented counterparts. $k$ decay is when weight decay is also applied to the $k$ parameters in an augmented network.}
    \label{cifar_comp}   
\end{table}

Table \ref{cifar_comp} shows the test error for two architectures: a ResNet with $n = 5$, and a Wide ResNet with $n = 4$, $n = 10$. Augmenting each model adds 15 and 12 parameters, respectively. We observe that $k$ decay hurts performance in both cases, indicating that they should either remain unregularized or suffer a more subtle regularization compared to the weight parameters. Due to its direct connection to layer degeneration, regularizing $k$ results in enforcing identity mappings, which might harm the model.

\begin{figure}[!ht]
  \centering
    \includegraphics[width=0.6\textwidth]{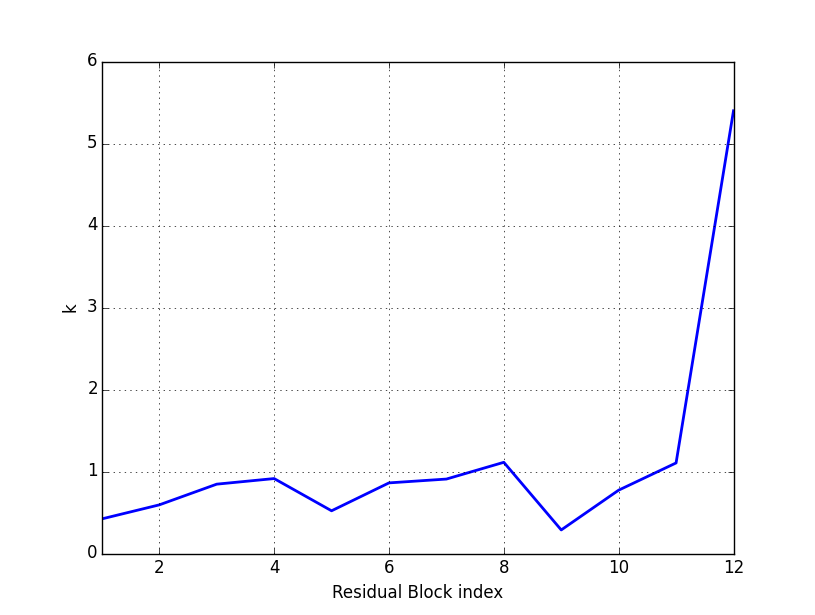}
  \caption{Values for $k$ according to ascending order of residual blocks. The first block, consisted of the first two layers of the network, has index 1, while the last block -- right before the softmax layer -- has index 12.}
\label{cpruning}
\end{figure}


As in the previous experiment, in Figure \ref{cpruning} we present the final $k$ values for each block. We can observe that the $k$ values follow an intriguing pattern: the lowest values are for the blocks of index $1$, $5$ and $9$, which are exactly the ones that increase the feature map dimension. This indicates that, in such residual blocks, the convolution performed in the shortcut connection to increase dimension is more important than the residual block itself. Additionally, the peak value for the last residual block suggests that its shortcut connection is of little importance, and could as well be fully removed without greatly impacting the model.


\begin{table}[h!]
\centering
    \begin{tabular}{ | l | c | c | c |}
    \hline
    Method    			& 		  	Params & C10+  & C100+  \\ \hline

    Network in Network (\cite{nin})			& 			-  & 	8.81  &  - \\ 
    FitNet (\cite{fitnet})			& 			-  & 	8.39   & 	35.04\\ 
    Highway Neural Network (\cite{highway})			& 			2.3M  & 	7.76   & 	32.39 \\
    All-CNN (\cite{allcnn})			& 			-  & 	7.25  & 	33.71  \\
\hline
    ResNet-110 (\cite{resnet1})			& 			1.7M  & 	6.61  & 	- \\ 
    ResNet in ResNet (\cite{rir})			& 			1.7M  & 	5.01   & 	22.90 \\ 
    Stochastic Depth (\cite{stdepth})	& 			10.2M & 	4.91   & 	- \\ 
    ResNet-1001 (\cite{resnet2}) & 			10.2M & 	4.62   & 	22.71 \\ 
    FractalNet (\cite{fractal}) & 			38.6M & 	4.60     & 	23.73 \\ 
    Wide ResNet (4,10) (\cite{wide})	& 			36.5M & 	3.89  & 	18.85 \\ 
    DenseNet (\cite{densenet})	& 			27.2M & 	3.74      & 	19.25\\ 
    Wide GatedResNet (4,10) + Dropout	& 			36.5M 	& 	\textbf{3.65}  & 	\textbf{18.27} \\ \hline
    \end{tabular}
\caption{Test error (\%) on the CIFAR-10 and CIFAR-100 dataset. All results are with standard data augmentation (crops and flips).}
    \label{cifar_all}   
\end{table}

Results of different models on the CIFAR datasets are shown in Table \ref{cifar_all}. The training and test errors are presented in Figure \ref{cifar}. To the authors' knowledge, those are the best results on CIFAR-10 and CIFAR-100 with moderate data augmentation -- only random flips and translations.

\section{Conclusion}

We have proposed a novel layer design based on Highway Neural Networks, which can be applied to provide general layers a quick way to learn identity mappings. Unlike Highway or Residual Networks, layers generated by our technique require optimizing only one parameter to degenerate into identity. By designing our method such that randomly initialized parameter sets are always close to identity mappings, our design offers less issues with optimization issues caused by depth.

We have shown that applying our technique to ResNets yield a model that can regulate the residuals, named Gated Residual Networks. This model performed better in all our experiments with negligible extra training time and parameters. Lastly, we have shown how it can be used for layer pruning, effectively removing large numbers of parameters from a network without necessarily harming its performance.

\bibliography{iclr2017_conference}
\bibliographystyle{iclr2017_conference}

\end{document}